\documentclass[letterpaper, 10pt, conference]{ieeeconf} 
\usepackage{FG2020}

\FGfinalcopy

\IEEEoverridecommandlockouts                             
\usepackage{tabularx, booktabs, ragged2e}
\usepackage[acronym]{glossaries}
\usepackage{acronym}
\usepackage{adjustbox}
\usepackage{multirow}
\usepackage{multicol}
\usepackage{balance}
\usepackage[colorlinks, linkcolor=red, urlcolor=blue]{hyperref}
\usepackage{amssymb}

\acrodef{fg}[FG]{15$^{th}$ IEEE International Conference on Automatic Face and Gesture Recognition}

\newcommand{\ie}{\textit{i}.\textit{e}., }
\newcommand{\eg}{\textit{e}.\textit{g}., }
\acrodef{gnn}[GNN]{graphical neural network}

\acrodef{ml}[ML]{machine learning}
\acrodef{sota}[SOTA]{state-of-the-art}

\acrodef{hog}[HOG]{histogram of gradients}

\acrodef{ss}[SS]{sister-sister}
\acrodef{bb}[BB]{brother-brother}
\acrodef{sibs}[SIBS]{brother-sister}

\acrodef{fs}[FS]{father-son}
\acrodef{ms}[MS]{mother-son}
\acrodef{fd}[FD]{father-daughter}

\acrodef{md}[MD]{mother-daughter}
\acrodef{gfgs}[GFGS]{grandfather-grandson}
\acrodef{gmgs}[GMGS]{grandmother-grandson}
\acrodef{gfgd}[GFGD]{grandfather-granddaughter}

\acrodef{gmgd}[GMGD]{grandmother-granddaughter}

\acrodef{sdm}[SDM]{signal detection model}
\acrodef{roc}[ROC]{receiver operating characteristic}
\acrodef{nmse}[NMSE]{Normalized Mean Square Error}
\acrodef{det}[DET]{Detection Error Trade-off}
\acrodef{tp}[TP]{true-positive}
\acrodef{tn}[TN]{true-negative}
\acrodef{ap}[AP]{average precision}

\acrodef{tpir}[TPIR]{true-positive identification rate}
\acrodef{frir}[FRIR]{false-reject identification rate}
\acrodef{fpir}[FRIR]{false-positive identification rate}

\acrodef{fn}[FN]{false-negative}
\acrodef{frr}[FRR]{false-reject rate}
\acrodef{fnr}[FNR]{false-negative rate}

\acrodef{fpr}[FPR]{false-positive rate}
\acrodef{tpr}[TPR]{true-positive rate}

\acrodef{fiw}[FIW]{\textit{Families In the Wild}}
\acrodef{tsk}[TSKIN]{\textit{Tri-Subject Kinship}}

\acrodef{kfw}[KinFaceW]{\textit{Kin-Faces in the Wild}}
\acrodef{kfvw}[KFVW]{\textit{KinFaceW Videos}}

\acrodef{rfiw}[RFIW]{\textit{Recognizing Families In the Wild}}

\acrodef{cnn}[CNN]{Convolutional Neural Network}
\acrodef{lut}[LUT]{Look-Up-Table}
\acrodef{fr}[FR]{face recognition}

\acrodef{svm}[SVM]{Support Vector Machine}
\acrodef{mid}[MID]{Member ID}
\acrodef{fid}[FID]{Family ID}
\acrodef{pid}[PID]{Photo ID}
\acrodef{roc}[ROC]{receiver operating characteristic}
\acrodef{nrml}[NRML]{Neighborhood Repulsed Metric Learning}

\definecolor{ao(english)}{rgb}{0.0, 0.5, 0.0}
	
\title{\LARGE \bf
\ac{rfiw}: The 4th Edition
}

\author{
    \parbox{16cm}{\centering
    \large Joseph P. Robinson$^1$, Yu Yin$^1$, Zaid Khan$^1$, Ming Shao$^2$,  Siyu Xia$^3$, Michael Stopa$^4$, Samson Timoner$^5$, Matthew A. Turk$^6$, Rama Chellappa$^7$, and Yun Fu$^1$\\
    \normalsize
    $^1$Northeastern University $^2$UMass Dartmouth $^3$Southeast University $^4$Konica Minolta \\
   $^5$ISM Connect $^6$Toyota Technological Institute at Chicago (TTIC) $^7$University of Maryland
}
}

\begin{document}

\ifFGfinal
\thispagestyle{empty}
\pagestyle{empty}
\else
\author{Anonymous FG2020 submission\\ Paper ID \FGPaperID \\}
\pagestyle{plain}
\fi
\maketitle

\acresetall

%%%%%%%%%%%%%%%%%%%%%%%%%%%%%%%%%%%%%%%%%%%%%%%%%%%%%%%%%%%%%%%%%%%%%%%%%%%%%%%%
\begin{abstract}
\ac{rfiw}-- an annual large-scale, multi-track automatic kinship recognition challenge supporting visual kin-based problems on scales larger than before. Organized in conjunction with the \ac{fg}, \ac{rfiw} provides a platform for publishing original work and the gathering of experts for a discussion of the next steps. This paper summarizes the supported tasks (\ie kinship verification, tri-subject verification, and search \& retrieval of missing children) in the evaluation protocols, which include the practical motivation, technical background, data splits, metrics, and benchmark results. Furthermore, top submissions (\ie leader-board stats) are listed and reviewed as a high-level analysis on the state of the problem. In the end, the purpose of this paper is to describe the 2020 \ac{rfiw} challenge, end-to-end, along with forecasts in promising future directions.

\end{abstract}
\hypersetup{citecolor=blue}
\acresetall
\glsresetall
%%%%%%%%%%%%%%%%%%%%%%%%%%%%%%%%%%%%%%%%%%%%%%%%%%%%%%%%%%%%%%%%%%%%%%%%%%%%%%%%
\section{Introduction}
Automatic kinship recognition has numerous uses. For instance - as an aid in forensic investigations, automated photo library management, historical lineage and genealogical studies, social-media-based analysis, tragedies of missing children and human trafficking, and concerns about immigration and border patrol. Nonetheless, the challenges in such face-based tasks (\ie fine-grained classification in unconstrained settings), are only amplified in the kin-based problem sets, as the data exhibits a high degree of variability in pose, illumination, background, and clarity,  along with soft bio-metric target labels (\ie kinship), which only further exacerbates the challenges with consideration for the directional relationships. Hence, the usefulness brought by the  practical benefits of enhancing kinship-based technology is matched by the challenges posed by the problem of automatic kinship understanding. This motivated the launching of the \ac{rfiw} challenge series: a large-scale data challenge in support of multiple tasks with the aim to advance kinship recognition technologies. We intend for \ac{rfiw} to serve as a platform for expert and junior researchers to present and share thoughts in an open forum.

The \ac{fiw} dataset~\cite{robinson2016families, robinson2018visual, wang2017kinship}-- a large-scale, multi-task image set for kinship recognition-- supports the annual \ac{rfiw}.\footnote{\ac{fiw} project page, \href{https://web.northeastern.edu/smilelab/fiw/}{https://web.northeastern.edu/smilelab/fiw/}.} The aim of the \ac{rfiw} challenge is to bridge the gap between research-and-reality using its large scale, variation, and rich label information. This makes modern-day data-driven approaches possible, as has been seen since its release in 2016~\cite{AdvNet, ertugrul2017will, gao2019will, li2017kinnet, wu2018kinship}.

We summarize the evaluation protocols-- practical motivation, technical background, data splits, metrics, and benchmarks-- of the 2020 \ac{rfiw} challenge. Specifically, this manuscript serves as a white-paper of the \ac{rfiw} held in conjunction with the \ac{fg}. Additional and information supplemental on the challenge website.\footnote{\ac{rfiw}2020 webpage, \href{https://web.northeastern.edu/smilelab/rfiw2020/}{https://web.northeastern.edu/smilelab/rfiw2020/}.} 

The remainder of the paper is organized as follows. The three tasks that make-up \ac{rfiw}2020 are introduced separately (Section \ref{sec:kinver}, \ref{sec:trisubject}, and \ref{sec:search}). For each task, a clear problem statement, the intended use, data splits, task protocols (\ie evaluation settings and metrics), and benchmark results are provided. From there, we bring up the discussion (Section~\ref{sec:discussion}) on broader impacts and potential next steps. Then, we conclude (Section~\ref{sec:conclusion}).
\begin{figure}[t!]
    \centering
    \includegraphics[width =.9\linewidth]{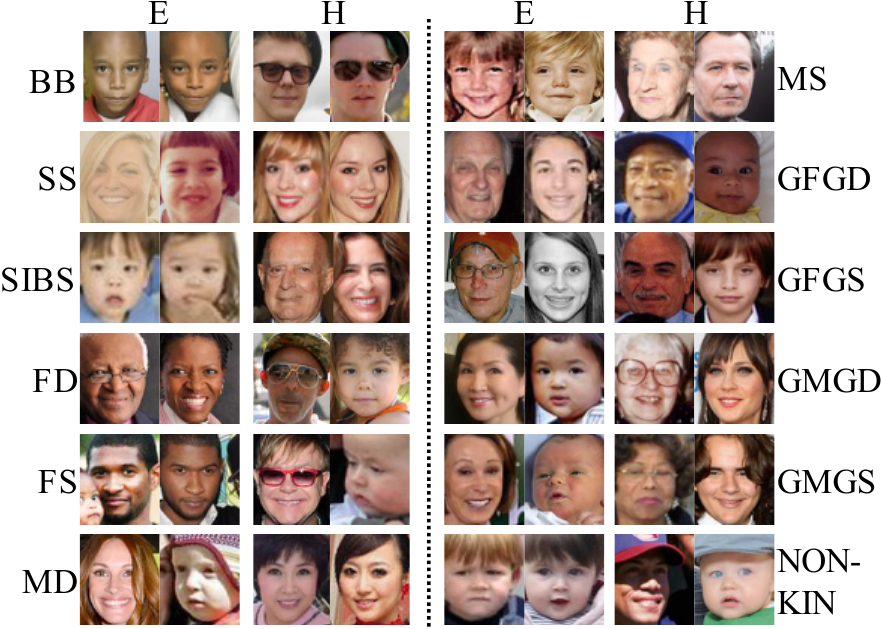}
    \caption{Sample pairs for the categories of T-1, kinship verification. For each, sample pairs with similarity scores near the threshold (\ie hard (H) samples), along with highly confident predictions (\ie easy (E) samples).}
    \label{fig:track1:samples}
\end{figure}

%% Multi-PIE setting 2
\begin{table*}[t!]
    \centering
   
    \caption{Counts for T-1: number of unique pairs (\textbf{P}), families (\textbf{F}), and face samples (\textbf{S}), with an increase in counts and types since~\cite{robinson2017recognizing}.}
    \scriptsize
    \begin{adjustbox}{width=\linewidth}
    \begin{tabular}{p{.1in}m{.1in}m{.29in}m{.29in}m{.29in}m{.29in}m{.29in}m{.29in}m{.29in}m{.29in}m{.29in}m{.29in}m{.29in}|m{.29in}}
    % \toprule
    & & BB& SS&SIBS& FD &FS & MD &MS & GFGD & GFGS &GMGD & GMGS& Total\\\hline
     \parbox[t]{2mm}{\multirow{3}{*}{\rotatebox[origin=c]{90}{Train}}}&\textbf{P} & 991  & 1,029 &1,588 & 712 & 721& 736& 716 & 136 & 124 & 116 & 114 &6,983\\
    % \multirow{3}{*}{Train}&\textbf{P} &  &  & &&&&&&&&\\
    \multirow{3}{*}{} &\textbf{F}  &303 & 304 & 286 & 401 & 404 & 399 & 402 & 81 & 73&71 & 66 &2790\\
    \multirow{3}{*}{} &\textbf{S} &39,608& 27,844 & 35,337& 30,746  &46,583 & 29,778&  46,969& 2,003 &  2,097  &1,741 & 1,834  &264,540\\\hline
    
    \parbox[t]{2mm}{\multirow{3}{*}{\rotatebox[origin=c]{90}{val}}} &\textbf{P}  & 433 & 433 & 206& 220 & 261 & 200 & 234 & 53 & 48 & 56 & 42 & 2,186 \\
    
    \multirow{3}{*}{} &\textbf{F}  &74  & 57& 90 & 134& 135& 124& 130& 32& 29& 36&27 &868\\
    \multirow{3}{*}{} &\textbf{S}  & 8,340 & 5,982 & 21,204& 7,575 &9,399&8,441 &7,587 & 762 &879 & 714 & 701 & 71,584\\\hline

    \parbox[t]{2mm}{\multirow{3}{*}{\rotatebox[origin=c]{90}{test}}} &\textbf{P}  &  469& 469 & 217 & 202& 257 & 230 & 237 & 40 & 31 & 36 & 33&2,221 \\
    \multirow{3}{*}{} &\textbf{F}  & 149  & 150  & 89 & 126 & 133 & 136 & 132 & 22 & 21 & 20 & 22 & 1,190\\
    \multirow{3}{*}{} &\textbf{S}  & 3,459 &2,956 &967 &3,019&3,273&3,184& 2,660 &121&96&71&84&39,743\\\hline
    
    \end{tabular}\label{tbl:track1:counts} 
    \end{adjustbox}
    \vspace{-5mm}
\end{table*}

\section{Related Works}
% \subsection{Kinship Recognition}
Kinship recognition, as seen in the machine vision, started in~\cite{fang2010towards}, where minimal data and low-level features set the stage for the task of kinship verification between parents and child. Soon thereafter, \cite{xia2012understanding} took a gender specific view of the problem-- moreover, the problem was viewed as a low rank transfer subspace problem, where the source and target are set as faces of the parent at younger and older ages, respectively~\cite{shao2012low}. Family101~\cite{fang2013kinship} was the first facial image dataset with family tree labels; at about the same time, KinWild~\cite{lu2014neighborhood} was released and used to organize data challenges~\cite{lu2015fg}. The task of tri-subject kinship verification (\ie Track 2), was inspired by the work that came next, in~\cite{qin2015tri}, for which data (\ie TS-Kin) and benchmarks were released. Until the release of \ac{fiw} in 2016~\cite{robinson2016families}, deep learning models were not widely applied to the kin-based domain, with the minimal exception (\ie \cite{zhang12kinship}), as the data capacity of their more complex machinery was not met by previous datasets. 
As part of the first \ac{rfiw}~\cite{robinson2017recognizing}), 
\ac{fiw} was further extended~\cite{robinson2018visual, wang2017kinship}, making ever more kin-based problems possible to approach~\cite{gao2019will, mingaaai2020}. A major focus of this (\ie \ac{rfiw} 2020) is to establish a record of state-of-the-art for the latest-and-greatest version of the \ac{fiw} image-set.

\section{Task Evaluations, Protocols, Benchmarks}\label{sec:task:eval}
\ac{rfiw} 2020 supported three tasks: kinship verification (T-1), tri-subject verification (T-2), and search \& retrieval of family members for missing children (T-3). We next describe each task separately, following the same outline: the problem statement and motivation, data splits and protocols, and benchmark experiments (\ie baselines). A brief section on experimental settings common to all tasks precedes the detailed descriptions of each task in separate subsections.

\subsection{Experimental settings}
The \ac{fiw} dataset provides the most extensive set of face pairs for kin-based face recognition. \ac{fiw} provides the data needed to train modern-day data-driven deep models~\cite{duan2017advnet, li2017kinnet, wang2017kinship, wu2018kinship}. \ac{fiw} was split into three parts:  \emph{train}, \emph{val}, and \emph{test}. Specifically, 60\% of the families were assigned to the  \emph{train} set; the remaining 40\% was split evenly between \emph{val} and \emph{test}. The three sets are completely disjoint in family and identity. Labeled \emph{train} and unlabeled \emph{val} were first released, with servers open for scoring (\emph{Phase 1}). Then, ground-truth for {\emph val} was made available (\emph{Phase 2}). Finally, the ``blind'' \emph{test} set was released at the start of \emph{Phase 3}. \emph{Phase 3} lasted for ten days to allow teams to process and make final submissions for scoring. Teams were asked to only process the \emph{test} set when generating submissions and any attempt to analyze or understand the \emph{test} pairs was prohibited.

\renewcommand{\arraystretch}{1.14}
\setlength{\arrayrulewidth}{0.4pt}

\begin{table*}[t]
\scriptsize
\centering
\caption {Averaged verification accuracy scores for T-1 of \ac{rfiw}.}
\label{tab:benchmark:track1}
\begin{tabular}{r|cccc|cccc|ccc|c}
  Methods & FD & FS & MD & MS & GFGD & GFGS & GMGD & GMGS & BB & SS & SIBS & Avg. \\
  \midrule
  Sphereface~\cite{Liu_2017_CVPR} (baseline) & 0.61 & 0.66 & 0.69 & 0.62 & 0.66 &0.71& 0.73 & \textbf{0.68} & 0.57 & 0.64 & 0.50 & 0.64\\
    stefhoer~\cite{id2} & \textbf{0.77} & 0.80 & 0.77 & \textbf{0.78} & 0.70 & \textbf{0.73} & 0.64 & 0.60 & 0.66 & 0.65 & 0.76 & 0.74\\
     ustc-nelslip~\cite{id6} & 0.76 & 0.82 & 0.75 & 0.75 & \textbf{0.79} & 0.69 & \textbf{0.76} & 0.67 & 0.75 & 0.74 & 0.72 & 0.76\\
     DeepBlueAI~\cite{id3} & 0.74 & 0.81 & 0.75 & 0.74 & 0.72 & \textbf{0.73} & 0.67 & \textbf{0.68} & 0.77 & 0.77 & 0.75 & 0.76\\
  vuvko~\cite{id4}& 0.75 & \textbf{0.81} & \textbf{0.78} & 0.74 & 0.78 & 0.69 & \textbf{0.76} & 0.60 & \textbf{0.80} & \textbf{0.80} & \textbf{0.77} & \textbf{0.78}\\\vspace{-5mm}
\end{tabular}
\end{table*}

As part of pre-processing, faces for all three sets were encoded via Sphereface \ac{cnn}~\cite{Liu_2017_CVPR}  (\ie 512 D). All pre-processing and the model weights were from the original work.\footnote{\href{https://github.com/wy1iu/sphereface}{https://github.com/wy1iu/sphereface}} Also common, is the use of cosine similarity to determine closeness of a pair of facial features $p_1$ and $p_2$~\cite{nguyen2010cosine}. This is defined as
$$
CS(\pmb p_1, \pmb p_2) = \frac {\pmb p_1 \cdot \pmb p_2}{||\pmb p_1|| \cdot ||\pmb p_2||}.
$$

Scores were then compared to threshold $\gamma$ (\ie $\text{score} > \gamma$ infers KIN; else, NON-KIN) or sorted (\ie T-3).

Scores were then either compared to threshold $\gamma$ (\ie $\text{cossim}(p_1, p_2) > \gamma$ infers KIN; else, NON-KIN) or sorted (\ie to rank in T-3). This concludes experimental settings common to all tasks.

\subsection{Kinship Verification}\label{sec:kinver}

Kinship verification aims to determine whether a pair of faces are blood relatives. This classical Boolean problem has two possible outcomes, KIN or NON-KIN (\ie true or false, respectively). Hence, this is the \textit{one-to-one} view of kin-based problems. The classical problem can be further extended by considering the type of kin relation between a pair of faces, rather than treating all kin relations equally~\cite{robinson2018recognize}.

Prior research mainly considered parent-child kinship types, \ie \ac{fd}, \ac{fs}, \ac{md}, \ac{ms}. Less attention has been given to sibling pairs, \ie \ac{ss}, \ac{bb}, and \ac{sibs}. Research findings in psychology and computer vision found that different relationship types share different familial features~\cite{Ming_CVPR11_Genealogical}. Hence, each relationship type can be modeled and evaluated independently. Thus, additional kinship types would further both our understanding and capabilities of automatic kinship recognition. With \ac{fiw}, the number of facial pairs accessible for kinship verification has dramatically increased, with a subset of the pair types and face pairs listed in Table~\ref{tbl:track1:counts}. Additionally, benchmarks now include grandparent-grandchildren types, \ie \ac{gfgd}, \ac{gfgs}, \ac{gmgd}, \ac{gmgs}.

%%%%%%%%%%%%%%%%%%%%%%%%%%%%%%%%%%%%%%%%%%%%%%%%%%%%%%%%%%%%%%%%%%%%%%%%%%%%%%%%
\subsubsection{Data Splits}
 \ac{fiw} supports eleven different relationship types that were used in \ac{rfiw} (Table~\ref{tbl:track1:counts}). The {\emph test} set had an equal number of positive and negative pairs and with no family (and, hence, subject identity) overlap between sets.

\subsubsection{Settings and metrics}\label{subsec:track1:settings}
Conventional face verification protocols were followed~\cite{LFWTech}, offering different modes (or settings) to span multiple paradigms of kinship verification. We next list the modes:
\begin{enumerate}
    \item \textit{Unsupervised:} No labels provided, \ie the prior knowledge about kinship or subject IDs.
    \item \textit{Image-restricted:} Kinship labels (\ie KIN/NON-KIN) will be provided for a training set that is completely disjoint from "blind" evaluation set, \ie no subject or family overlap between training and evaluation sets.
    \item \textit{Image unrestricted:} Along with the kinship labels, subject IDs are provided. This allows for the ability to generate additional negative pair-wise samples.
\end{enumerate}

Verification accuracy is used to evaluate. Specifically,

$$
\text{Acc.}_j = \frac{\text{\# correct predictions for j-th type}}{\text{Total \# of pairs for j-th type}},
$$
where $j^{th}\in\{\text{all 11 relationship types}\}$. Then, the the overall accuracy is calculated as a weighted sum (\ie weight by the pair count to determine the average accuracy).

\subsubsection{Baseline Experiments}
The threshold was determined by the value that maximizes the accuracy on the  \emph{val} set. Results are listed in Table~\ref{tab:benchmark:track1}, with samples in Fig~\ref{fig:track1:samples}.

%%%%%%%%%%%%%%%%%%%%%%%%%%%%%%%%%%%%%%%%%%%%%%%%%%%%%%%%%%%%%%%%%%%%%%%%%%%%%%%%
\subsection{Tri-Subject Verification}\label{sec:trisubject}
% \subsection{Problem statement and intended use}
Tri-Subject Verification focuses on a different view of kinship verification-- the goal is to decide if a child is related to a pair of parents. First introduced in~\cite{qin2015tri},  it makes a more realistic assumption, as having knowledge of one parent often means the other potential parent(s) can be easily inferred.

Triplet pairs consist of Father ({F}) / Mother ({M}) - Child ({C}) ({FMC}) pairs, where the child {C} could be either a Son ({S}) or a Daughter ({D}) (\ie triplet pairs are {FMS} and {FMD}).

\begin{table}[b]
    \centering
    
    \caption{Counts for T-2. No. of pairs (\textbf{P}), families (\textbf{F}), face samples (\textbf{S}).}
    % \scriptsize
    %\begin{adjustbox}{width=\linewidth}
    \small
    \begin{tabular}{p{.1in}m{.1in}ccc}
    & &FM-S &FM-D &Total\\\hline
     \parbox[t]{2mm}{
     \multirow{3}{*}{\rotatebox[origin=c]{90}{train}}}&\textbf{P} & 662  & 639 &1,331 \\
    \multirow{3}{*}{} &\textbf{F}  &375 & 364 & 739\\
    \multirow{3}{*}{} &\textbf{S} &8,575& 8,588 &  17,163\\\hline
    
    \parbox[t]{2mm}{
    \multirow{3}{*}{\rotatebox[origin=c]{90}{val}}} &\textbf{P}  & 202 & 177 & 379 \\
    \multirow{3}{*}{} &\textbf{F}  &116  & 117& 233\\
    \multirow{3}{*}{} &\textbf{S}  & 2,859 & 2,493 & 5,352 \\\hline
    \parbox[t]{2mm}{
    \multirow{3}{*}{\rotatebox[origin=c]{90}{test}}} &\textbf{P}  &  205& 178 & 383  \\
    \multirow{3}{*}{} &\textbf{F}  & 116  & 114  & 230 \\
    \multirow{3}{*}{} &\textbf{S}  & 2,805 &2,400 &5,205\\\hline
    
    \end{tabular}\label{tbl:track2:counts} 
\end{table}

\subsubsection{Data Splits}

Following the procedure in \cite{qin2015tri}, we create positive (have kin relation) triplets by matching each husband-wife spouse pair with their biological children, and negative (no kin relation) triplets by shuffling the positive triplets until every spouse pair is matched with a child which is not theirs (Table~\ref{tbl:track2:counts}).
Because the number of potential negative samples far exceeds the number of potential positive examples, we only generate one negative triplet for each positive triplet, again following the procedure of \cite{qin2015tri}. 

We post-process the positive triplets before generating negatives to ensure balance among individuals, families, and spouse pairs, since a naive data selection procedure which weights every face sample similarly would result in some individuals and families being severely over-represented due to an abundance of face samples for some identities and families. 
The post-processing is done by limiting the number of samples of any triplet $(F, M, C)$, where $F$, $M$, and $C$ are identities of a father, mother, and child to 5, then limiting the appearance of each $(F, M)$ spouse-pair to 15, and then finally limiting the number of triplet samples from each family to 30. The \emph{test} set has an equal number of positive and negative pairs. Lastly, note that there is no family or subject identity overlapping between any of the sets.

\subsubsection{Settings and metrics}
Per convention in face verification, we offer 3 modes (\ie the same as in task 1 listed in Section~\ref{subsec:track1:settings}). The metric used is, again, verification accuracy, which is first calculated per triplet-pair type (\ie FMD and FMS). Then, the weighted sum (\ie average accuracy) determines the leader-board.

\subsubsection{Baseline Results}
Baseline results are shown in Table~\ref{tab:benchmark:track2}. A score was assigned to each triplet $(F_i, M_i, C_i)$ in the validation and \emph{test} sets using the formula $$ \text{score}_{i} =  avg(\cos{(F_i, C_i)}, \cos{(M_i, C_i)}) $$
where $F_i$, $M_i$ and $C_i$ are the feature vectors of the father, mother, and child images respectively from the i-th triplet. 
Scores were compared to a threshold $\gamma$ to infer a label (\ie predict KIN if the score was above the threshold; else, NON-KIN). 
The threshold was found experimentally on the \emph{val} set. The threshold was applied to the \emph{test} (Table~\ref{tab:benchmark:track2}).

\begin{table}[b]
\scriptsize
\centering
\caption {Tri-subject verification accuracy scores for T-II benchmark.}
\label{tab:benchmark:track2}
\begin{tabular}{r|cc|c}
  &FMS & FMD & Avg. \\
  \midrule
  
  Sphereface~\cite{Liu_2017_CVPR} (baseline) & 0.68 & 0.68 & 0.68 \\ 
    stefhoer~\cite{id2} & 0.74 & 0.72 & 0.73 \\
  DeepBlueAI~\cite{id3}  & 0.77 & 0.76 & 0.77 \\
 ustc-nelslip~\cite{id6}  & \textbf{0.80} & \textbf{0.78} & \textbf{0.79} \\
\end{tabular}
\end{table}

\begin{figure}[t!]
    \centering
    \includegraphics[width = .9\linewidth]{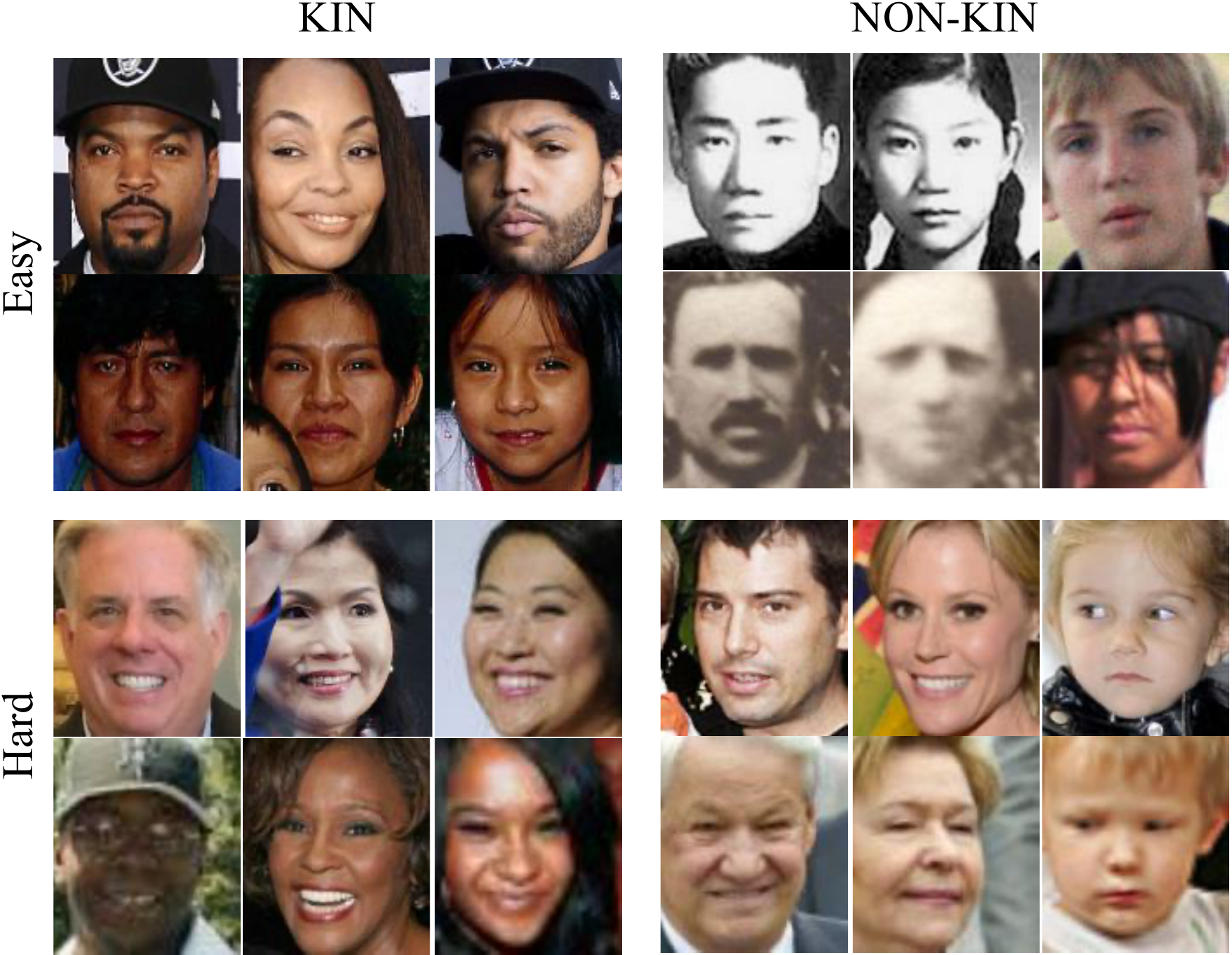}
    \caption{Tri-subject pairs near the threshold, and for correct and incorrect predictions. Each shows FMS (top rows) and FMD (bottom).}
    \label{fig:track2:montage}
\end{figure}

\subsection{Search and Retrieval}\label{sec:search}
T-3 is posed as a \textit{many-to-many}, \ie one-to-many samples per subject. Thus, we imitate template-based evaluations on the probe side, but faces in the gallery are not labeled by subject. Furthermore, the goal is to find relatives of search subjects (\ie \textit{probes}) in a search pool (\ie \textit{gallery}).

Kin information, as a search cue, can be leveraged to improve conventional FR search systems, or even as prior knowledge for mining social or family relationships in industries like \textit{Ancestry.com}. However, the task is most directly related to missing persons. Thus, we formulate it as such.

\begin{figure}[t!]
    \centering
    \includegraphics[width = .8\linewidth]{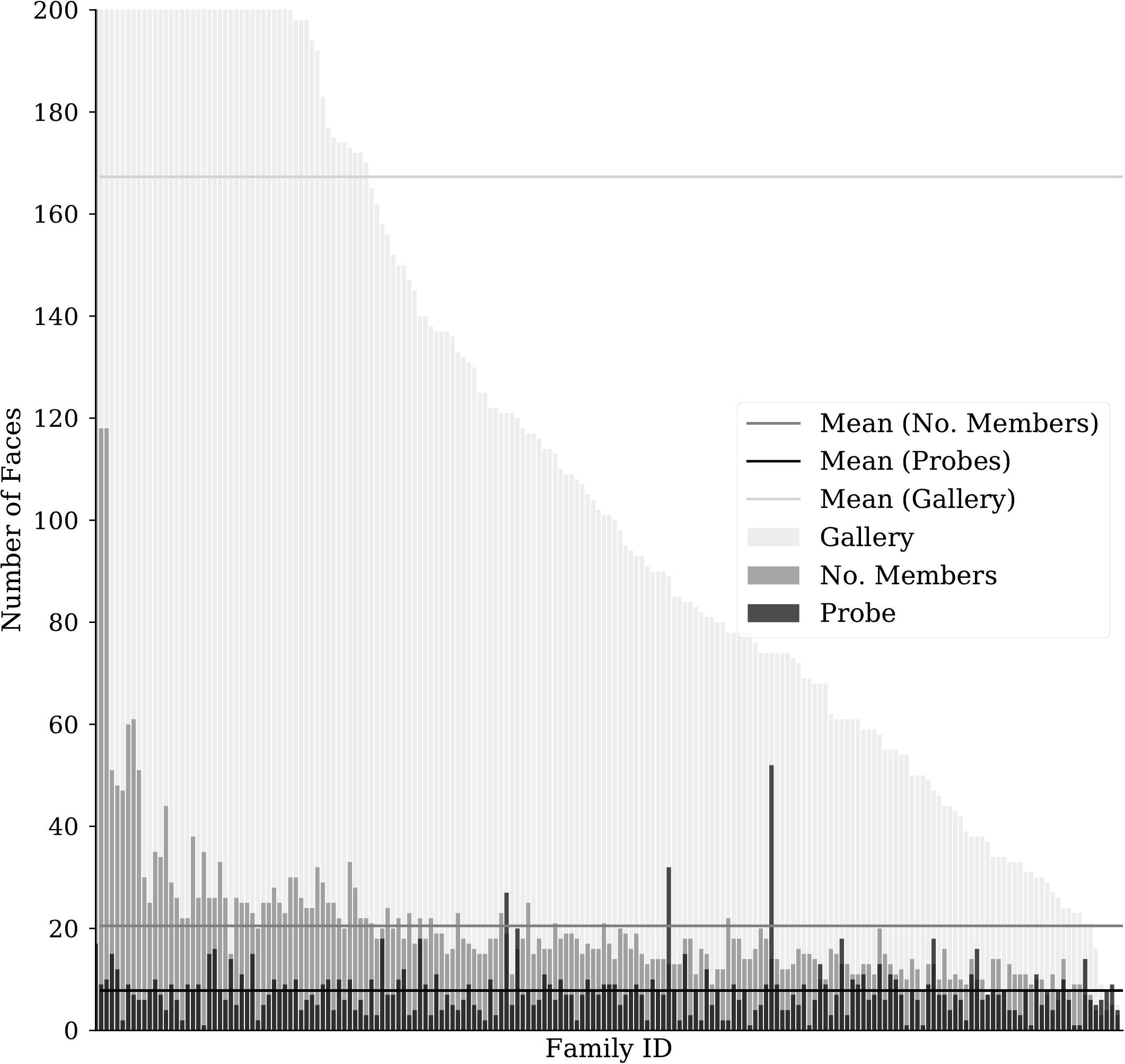}
    \caption{Plot showing the face counts for each family in {\emph test} set of T-3. The probes have about 8 faces on average, while the number of family members in the gallery nears 20 on average, with an average of 170 faces in total.}
    \label{fig:track3:counts}
\end{figure}

The protocol of T-3 could be used to find parents and other relatives of unknown, missing children. The gallery contains 31,787 facial images from 190 families (Fig.~\ref{fig:track3:counts}): inputs are subject labels (\ie probes), and outputs are ranked lists of all faces in the gallery. The number of relatives varies for each subject, ranging anywhere from 0 to 20+. Furthermore, probes have one-to-many samples-- the means of fusing samples of probes is an open research question. This \textit{many-to-many} task is currently setup in closed form (\ie every probe has relative(s) in gallery).

\subsubsection{Data Spits}
This task will be composed of search subjects (\ie \textit{probes}) from different families. \textit{Probes} are supported by several samples of query subject, text description of family (\eg ethnicity, some relationships between selected members, etc.), and list of relatives present in \textit{gallery}. The \textit{test} set will only consist of sets of images for the probes. Diversity in terms of ethnicity is ensured for both sets. Again, three disjoint sets were split (Table~\ref{tbl:track3:counts}).

\begin{table}[b]
    \centering
    \caption{Counts for T-3: individuals (\textbf{I}), families (\textbf{F}), face samples (\textbf{S}).}
    \begin{tabular}{p{.1in}m{.1in}ccc}
    & &Probe &Gallery &Total\\\hline
     \parbox[t]{2mm}{
     \multirow{3}{*}{\rotatebox[origin=c]{90}{train}}}&\textbf{I} & --  & 3,021 & 3,021 \\
    \multirow{3}{*}{} &\textbf{F}  &-- & 571 & 571\\
    \multirow{3}{*}{} &\textbf{S} & --& 15,845 & 15,845 \\\hline
    
    \parbox[t]{2mm}{
    \multirow{3}{*}{\rotatebox[origin=c]{90}{val}}} &\textbf{I}  & 192 & 802  & 994  \\
    \multirow{3}{*}{} &\textbf{F} & 192 & 192 & 192  \\
    \multirow{3}{*}{} &\textbf{S}  &1,086  &4,030 &5,116 \\\hline

    \parbox[t]{2mm}{
    \multirow{3}{*}{\rotatebox[origin=c]{90}{test}}} &\textbf{I}& 190 & 783  & 9d73 \\
    \multirow{3}{*}{} &\textbf{F} &190  & 190  & 190   \\
    \multirow{3}{*}{} &\textbf{S}  &1,487  & 31,787 & 33,274\\\hline
    
    \end{tabular}\label{tbl:track3:counts} 
\end{table}

\subsubsection{Evaluation Settings} 
Each subject (\ie probe) gets searched independently, with 190 in total: hence, 190 families make-up the \textit{test} set. Probes have one-to-many faces. Following template conventions of other \textit{many-to-many} face evaluations, facial images for unique subjects are separated by identity, with a gallery containing variable number of relatives, each with a variable number of faces~\cite{whitelam2017iarpa}.

Teams were allowed to submit up to six final submissions, with each submissions being a ranked-list of all subjects in the gallery. Submissions were accompanied by a brief (text) description of the system used to generate results. With that was a ranked list per \textit{probe} in the \textit{test}. Per \ac{rfiw} rules, participants were permitted to analyze \emph{test} results, as this was the purpose of the 192 families provided as the \emph{val} set.

\paragraph{Evaluation Metric} 
MAP was the underlying metric used for comparisons. Mathematically speaking, scores for each of the $N$ missing children are calculated as follows:
$$AP(f)=\frac{1}{P_F}\sum^{P_F}_{tp=1}Prec(tp)=\frac{1}{P_F}\sum^{P_F}_{tp=1}\frac{tp}{rank(tp)}.$$
where average precision (AP) is a function of family $f$ with a total of ${P_F}$ \ac{tpr}. We then average all AP scores to determine overall MAP score as follows:
$$MAP = \frac{1}{N}\sum^{N}_{f=1}AP(f),$$

Additionally, \ac{tpr} as a function of rank will traced out for further analysis between different attempts.

\subsubsection{Baseline Results}
\begin{figure}[t!]
    \centering
    \includegraphics[width = .85\linewidth]{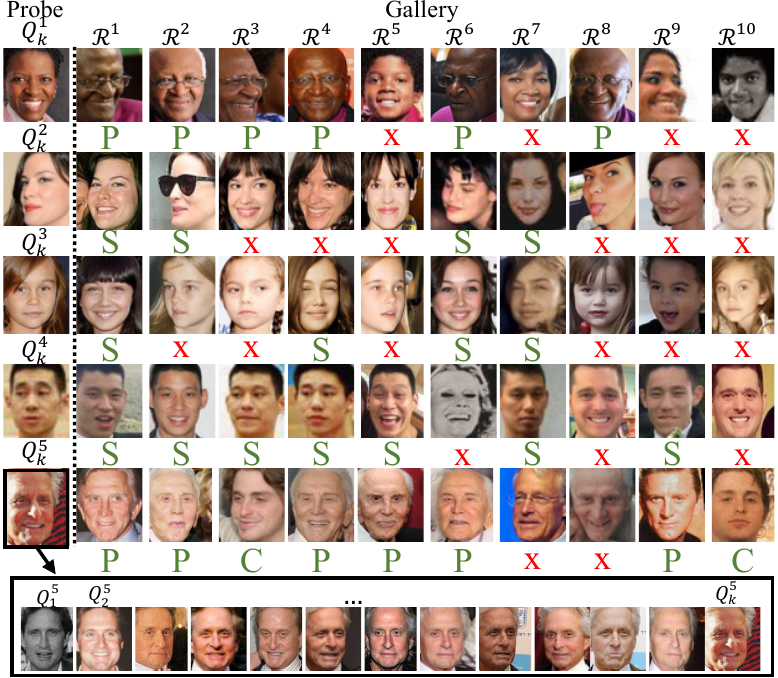}
    \caption{T-3 sample results (Rank 10). For each query (row) one or more faces of the probe returned the corresponding samples of gallery as top 10. Here, \textcolor{red}{x} (red) depicts false predictions, while true predictions displays the relationship type (in green): \textcolor{ao(english)}{P} for parent; \textcolor{ao(english)}{C} for child; \textcolor{ao(english)}{S} for sibling.}
    \label{fig:track3:montage}
    \vspace{-5mm}
\end{figure}

Table \ref{tbl:t3:benchmarks} and shown in Fig.~\ref{fig:track3:montage}.

\begin{table}[b!]
%\Large
	\centering
	\caption{Performance ratings for Track 3.}
	\begin{tabular}{c|cc} 
	      \textbf{Methods}  &\textbf{mAP} & \textbf{Rank@5} \\ \hline
		  Baseline (Sphereface)~\cite{Liu_2017_CVPR} & 0.02 & 0.10	\\
		  HCMUS notweeb~\cite{id9} & 0.07 & 0.28	\\
		  DeepBlueAI~\cite{id3} & 0.06 & 0.32	\\
		  ustc-nelslip~\cite{id8} & 0.08 & 0.38	\\
		  vuvko~\cite{id4} & \textbf{0.18} & \textbf{0.60}	\\
	\end{tabular}
	\label{tbl:t3:benchmarks}
\end{table}
%%%%%%%%%%%%%%%%%%%%%%%%%%%%%%%%%%%%%%%%%%%%%%%%%%%%%%%%%%%%%%%%%%%%%%%%%%%%%%%%
\section{Summary of submissions}

Solutions for the tasks of the 2020 \ac{rfiw} \ac{fg} challenge tended to use backbone networks trained for conventional \ac{fr}, then fine-tuned for kin-specific face tasks. Each submission for all three tasks surpasses the simple baseline provided as part of the challenge organization. We next summarize results of each team separately.

% \subsection{Kinship verification (T1)}
\subsection{Team Vuvko}
\emph{Team Vuvko}~\cite{id4} treated the different relationship types as a multi-task problem and trained a local expert for each type on top of a ResNet50~\cite{he2016deep}, simultaneously. This multi-task model, trained and evaluated for kinship verification (Table~\ref{tab:benchmark:track1}), was deployed for the other tasks as well (Table~\ref{tab:benchmark:track2} and~\ref{tbl:t3:benchmarks}). Another method applicable to all tasks was using different fusion techniques in deep feature space~\cite{id6, id8}. 

\begin{figure}[t!]
    \centering
    \includegraphics[width =.9\linewidth]{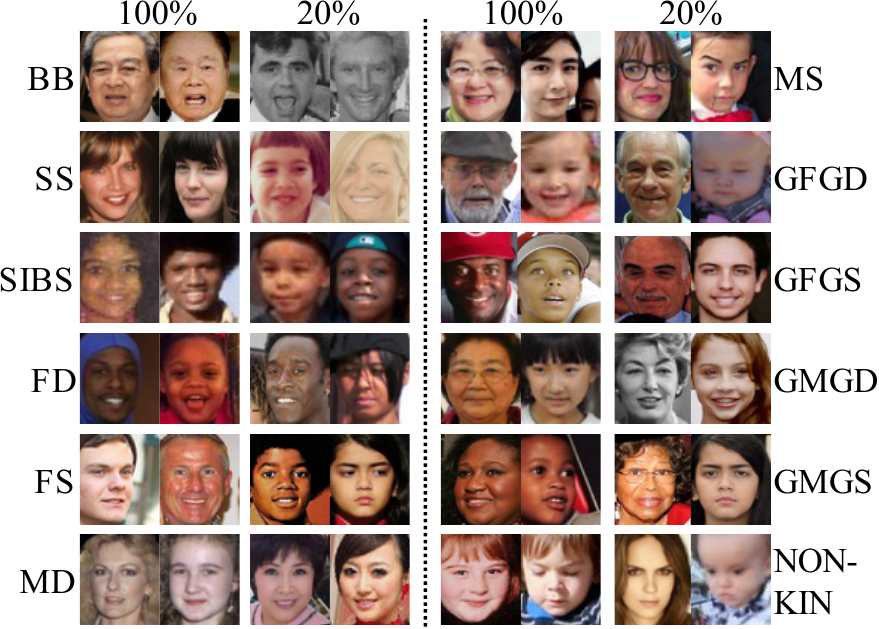}
    \caption{Sample pairs of T-1 that all teams got correct or incorrect.}
    \label{fig:track1:samples:submitted}
    \vspace{-6mm}
\end{figure}

\begin{figure}[t!]
    \centering
    \includegraphics[width =.6\linewidth]{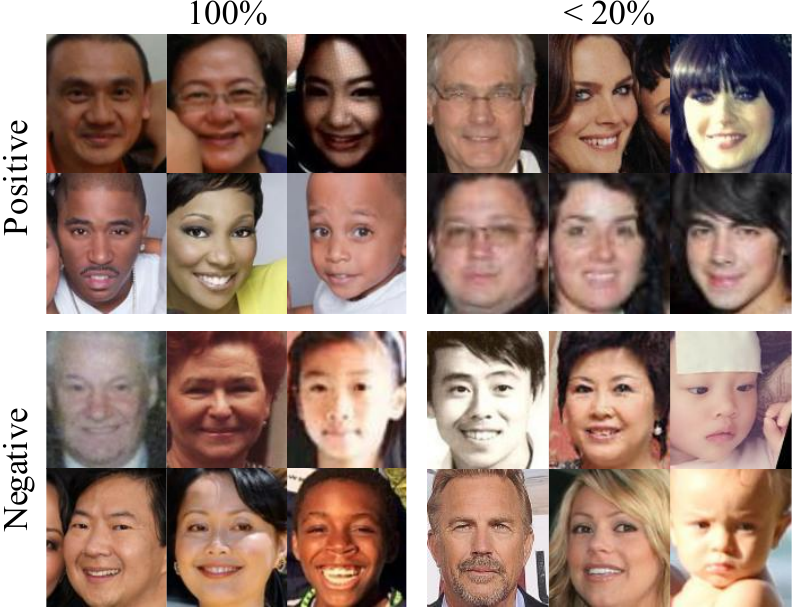}
    \caption{Sample triplets of T-2 that all teams got correct (left) or mostly incorrect (right). Each shows FMS (top rows) and FMD (bottom).}
    \label{fig:track2:samples:submitted}
       \vspace{-4mm}
\end{figure}

Sample pairs in the T-1 challenge that were unanimously correctly and incorrectly classified are shown in Fig.~\ref{fig:track1:samples:submitted}. Similarly, sample triplets that all teams got correct or incorrect in T-2 are shown (Fig.~\ref{fig:track2:samples:submitted}, left and right column, respectively).

\emph{Team Vuvko} scored the highest average in T-1 (Table~\ref{tab:benchmark:track1}) as well as the highest ranking for T-3 (Table~\ref{tab:benchmark:track2}).

\subsection{Team DeepBlueAI}
\emph{Team DeepBlueAI} used two pre-trained \acp{cnn} (\ie VGG-Face~\cite{schroff2015facenet} trained on VGG2~\cite{cao2018vggface2} and FaceNet trained on MSCeleb~\cite{guo2016ms})~\cite{id3}. The \acs{cnn} were used to encode each face-- the two face encodings were then concatenated using different types of arithmetic~\cite{id6, id8}. In \cite{id3}, the distance between faces was then determined using euclidean distance. Also, SENet~\cite{iandola2016squeezenet} was swapped in for ResNet50 as the backbone for a modest boost in performance on the validation, but dropped on the test. Much like in~\cite{robinson2018visual}, \cite{id3} fine-tuned a \ac{cnn} using families as the classes (\ie the difference was the authors used Arcface, opposed to Sphereface as in~\cite{robinson2018visual}).

\emph{Team DeepBlueAI} tied for second in kinship verification, T-1 (Table~\ref{tab:benchmark:track1}), and second to best in T-2 (Tablee~\ref{tab:benchmark:track2}).

\subsection{Team Ustc-Nelslip}
\emph{Team Ustc-nelslip}~\cite{id6} also used a Siamese network, i.e. encoding features from images in parallel with weights shared across the two image processings. ResNet50 or SENet50 was used as the backbone, both pre-trained on VGGFace2~\cite{cao2018vggface2}. In addition, team ustc-nelslip also employed two loss functions - binary cross-entropy and focal loss. Finally, they fused the feature vectors with two algebraic formulae leading to \( 2 \times 2 \times 2 = 8 \) independent "models." A unique feature was the construction of a "jury system" to combine outputs of different models to improve accuracy.

\emph{Team Ustc-nelslip} scored highest in T-2 (Table~\ref{tab:benchmark:track2}). 

\subsection{Team Stefhoer}
\emph{Team Stefhoer}~\cite{id2} placed particular emphasis on the the dependence of family identification accuracy for cross-gender versus same-gender pairs of images. These researchers constructed a Kinship \emph{comparator} module that consisted of eleven separate "local expert networks" connected in series. These eleven networks corresponded to the eleven types of family relationships (\eg father-son and brother-sister) in the challenge. Perhaps as a result of this focus team Stefhoer registered the highest score in the subcategories of father-daughter and mother-son identification (within T-1).

\subsection{Team HCMUS}
\emph{Team HCMUS}~\cite{id9} competed in Tracks I (kinship verification) and III (kinship search and retrieval). For extracting features the authors use a Siamese CNN with FaceNet (Inception-ResNet-v1) and with VGG-Face (Resnet-50) as the pre-trained models. FaceNet uses Triplet Loss as the main loss function in the training phase. The authors also implement ArcFace~\cite{deng2019arcface} - a family of loss functions based on the geodesic distance between feature vectors which aim to discriminate the latent representation of deep NNs.

\acresetall
\section{Discussion}\label{sec:discussion}
\subsection{A Broader Impact}
The fourth \ac{rfiw} gained fair attention. 
T-1, kinship verification, saw the most (10+ submissions). T-2 (\ie tri-subject) and T-3 (search and retrieval) were both supported for the first time by \ac{rfiw}, are more complex than the classic task of T-1, and are practically motivated. All submissions outscored baselines.

The scope of kin-based problems spans much wider than \ac{rfiw}. Specifically, in application (\eg generative-based tasks~\cite{gao2019will, ozkan2018kinshipgan}) and experimental settings~\cite{mingaaai2020}, focuses on particular views of the visual kinship recognition problem. Tasks of \ac{rfiw} were thought to be appropriate, provided the difficulty and practicality; the question how best to formulate the problem is an open research question, in itself.

\subsection{Conclusion}\label{sec:conclusion}
This paper presented the 2020 \ac{rfiw} challenge organized in conjunction with the \ac{fg}. The 2020 challenge is the fourth edition of the \ac{rfiw} annual evaluation. For this, we added 2 new tracks, tri-subject verification and search \& retrieval of missing children; the traditional kinship verification task continued to be supported as well. The \ac{fiw} dataset was used to pose each of the challenge tracks. As challenging it may be, many entries outperformed the ``vanilla'' baselines in all tasks. Regardless, in all three cases, there still exists much room for improvement. Accuracy on the Verification and Tri-subject has just begun to approach the 80\%; Search \& Retrieval further behind. Baseline code at \href{https://github.com/visionjo/pykinship}{github.com/visionjo/pykinship}. As we see it, the story of \ac{fiw} is still in its infancy.

{         
\bibliographystyle{ieee}

\scriptsize
\balance
\bibliography{rfiw2020} 

}
\end{document}